\documentclass{article}
\usepackage{ijcai25}

\usepackage{times}
\usepackage{soul}
\usepackage{url}
\usepackage[hidelinks]{hyperref}
\usepackage[utf8]{inputenc}
\usepackage[small]{caption}
\usepackage{graphicx}
\usepackage{amsmath}
\usepackage{amsthm}
\usepackage{booktabs}
\usepackage{algorithm}
\usepackage{algorithmic}
\usepackage[switch]{lineno}

\usepackage{amsxtra, amsfonts, amssymb, amstext, mathtools}
\usepackage{nicefrac}
\usepackage{xspace}
\usepackage{graphics,color}
\usepackage{wrapfig}
\usepackage{subcaption}
\usepackage{epstopdf}

\newtheorem{theorem}{Theorem}
\newtheorem{lemma}[theorem]{Lemma}

\newtheorem{definition}[theorem]{Definition}

\newcommand{\NSGA}{\mbox{NSGA}\nobreakdash-II\xspace}
\newcommand{\NSGAT}{\mbox{NSGA}\nobreakdash-III\xspace}
\newcommand{\SMS}{\mbox{SMS-EMOA}\xspace}

\newcommand{\oneminmax}{\textsc{OneMinMax}\xspace}
\newcommand{\omm}{\textsc{OneMinMax}\xspace}

\newcommand{\lotz}{\textsc{LOTZ}\xspace}

\newcommand{\mei}{\textsc{MEI}\xspace}
\newcommand{\start}{\rm{start}\xspace}
\newcommand{\opt}{\rm{opt}\xspace}

\newcommand{\R}{\ensuremath{\mathbb{R}}}

\newcommand{\N}{\ensuremath{\mathbb{N}}} 

\DeclareMathOperator{\nad}{nad}

\let\originalleft\left
\let\originalright\right
\renewcommand{\left}{\mathopen{}\mathclose\bgroup\originalleft}
\renewcommand{\right}{\aftergroup\egroup\originalright}

\title{The First Theoretical Approximation Guarantees for the Non-Dominated Sorting Genetic Algorithm III (NSGA-III)}

\author{
Renzhong Deng$^1$
\and
Weijie Zheng$^1$\footnote{Corresponding author.}\and
Benjamin Doerr$^{2}$\\
\affiliations
$^1$ School of Computer Science and Technology, National Key Laboratory of Smart Farm Technologies and Systems, International Research Institute for Artificial Intelligence, 
\\
Harbin Institute of Technology, Shenzhen\\
$^2$Laboratoire d'Informatique (LIX), CNRS, \'Ecole Polytechnique, \\
Institut Polytechnique de Paris, Palaiseau, France\\
\emails
\{dengrenzhong, zhengweijie\}@hit.edu.cn
}

\begin{document}

\maketitle
\sloppy{
\begin{abstract}
This work conducts a first theoretical analysis studying how well the \NSGAT approximates the Pareto front when the population size $N$ is less than the Pareto front size. We show that when $N$ is at least the number $N_r$ of reference points, then the approximation quality, measured by the maximum empty interval (\mei) indicator, on the \omm benchmark is such that there is no empty interval longer than $\lceil\frac{(5-2\sqrt2)n}{N_r-1}\rceil$. This bound is independent of $N$, which suggests that further increasing the population size does not increase the quality of approximation when $N_r$ is fixed. This is a notable difference to the \NSGA with sequential survival selection, where increasing the population size improves the quality of the approximations. 
We also prove two results indicating approximation difficulties when $N<N_r$. 
These theoretical results suggest that the best setting to approximate the Pareto front is $N_r=N$. In our experiments, we observe that with this setting the \NSGAT computes optimal approximations, very different from the \NSGA, for which optimal approximations have not been observed so far. 
\end{abstract}

\section{Introduction}

The non-dominated sorting genetic algorithm II (\NSGA)~\cite{DebPAM02} is the most widely used multi-objective evolutionary algorithm  (MOEA). The recent first mathematical runtime analysis of this algorithm~\cite{ZhengLD22,ZhengD23aij} has inspired many theoretical works on domination-based MOEAs such as the SPEA2~\cite{ZitzlerLT01}, the \SMS~\cite{BeumeNE07}, and the \NSGAT~\cite{DebJ14}. Notable results include \cite{BianQ22,DoerrQ23tec,DangOSS23aaai,BianZLQ23,DinotDHW23,WiethegerD23,ZhengD24many,ZhengD24,OprisDNS24,RenBLQ24,DoerrIK25}. 
Interestingly, these results suggest that the slightly less prominent algorithms SPEA2, \SMS, and \NSGAT are more powerful than the \NSGA, at least when the number of objectives is three or more. This is the reason why in this work we concentrate on one of them, namely the \NSGAT. 

Almost all of the existing runtime analyses regard the complexity of computing the full Pareto front. In practice, this is often not feasible, because the Pareto front is too large, and it may also not be desirable, since ultimately a human decision maker has to select one of the computed solutions as the solution to be adopted. For this reason, we shall discuss the approximation qualities of the \NSGAT in this work. 

In the first and only mathematical work discussing the approximation abilities of one of the above-named algorithms, Zheng and Doerr~\shortcite{ZhengD22gecco,ZhengD24approx} analyzed how well the \NSGA approximates the Pareto front of the \mbox{\omm} benchmark when its population size $N$ is less than the size of the Pareto front. 
They first observed that the classic \NSGA, which first computes the crowding distance and then, based on these numbers, selects the next population, can compute very bad approximations, creating empty intervals on the Pareto front by arbitrary factors larger than what an optimal approximation displays. 
This can be overcome by using the sequential \NSGA proposed in \cite{KukkonenD06}, which removes individuals sequentially, always updating the crowding distance values after each removal, or the steady-state \NSGA, which generates only one offspring per iteration and hence also removes only one individual per iteration. For these two variants of the \NSGA, it was proven that within an expected number of $O(Nn\log n)$ function evaluations, approximations with largest empty interval size $\mei\le \max\{ \frac{2n}{N-3}, 1\}$ are computed. Since the optimal $\mei$ value for the \omm benchmark and population size $N$ is $\lceil\frac{n}{N-1}\rceil$, this is essentially a $2$-approximation. No such results exist for any other of the domination-based algorithms, in particular, not for the \NSGAT.

\textbf{Our contributions:} To fill this research gap, this work will conduct the first analysis of the approximation ability of the \NSGAT, and will detect some notable differences to the \NSGA. Since the \omm problem is the only benchmark for which the approximation abilities of the \NSGA-type algorithms were studies, in this first approximation work for the \NSGAT we shall also regard this bi-objective problem. We are aware of the fact that generally the \NSGAT is seen as an algorithm for many-objective optimization, but in this first work our focus is on a comparison with the \NSGA via the results obtained in \cite{ZhengD24approx}, so for that reason we restrict ourselves to two objectives. From our proofs, we would conjecture that our findings can be generalized to more objectives. 

The main approximation guarantee we prove is that with a population size $N$ at least as large as the number $N_r$ of reference points, the \NSGAT computes approximations to the Pareto front of the \omm problem with 
$\mei \le \lceil\frac{(5-2\sqrt2)n}{N_r-1}\rceil$, and this within an expected number of $O(Nn^c\log n)$ function evaluations, where $c=\lceil\frac{2(2-\sqrt2)}{N_r-1}\rceil$. Recalling that the optimal $\mei$ is $\mei=\lceil\frac{n}{N-1}\rceil$, we see that also the \NSGAT can compute constant factor approximations. Our result shows this factor to be at most $5-2\sqrt 2 \approx 2.17$, slightly larger than the factor of $2$ shown for the sequential and steady-state \NSGA.

We also prove that when $N < N_r$, the approximation can be worse than an optimal one by a factor of $\Omega(\log n)$. These results suggest that the number of reference points is best set to be equal to the population size, that is, $N_r=N$. This is consistent with (and thus supports) the suggestion (without theoretical explanation) to take $N_r \approx N$ made in the original \NSGAT paper~\cite{DebJ14}. 

Experiments are conducted to see how the \NSGAT approximates the Pareto front for \omm. The results show that with $N_r=N$, the \NSGAT performs better than the \NSGA with sequential survival selection, and always reaches the optimal approximation. This observation suggests that proving a tighter approximation bound is an interesting target for future research. The experiments also consider the case where $N_r>N$, and show that the approximation ability of the \NSGAT becomes worse as $N_r$ increases, and even the  extremal points of the population can be lost (which cannot happen for the \NSGA as these points have infinite crowding distance). To verify the generalizability of the above theoretical findings and experimental observations on \omm, we also conduct  experiments for the popular \lotz benchmark. We again observe that the setting of $N_r=N$ results in optimal approximations, and that large numbers of reference points, that is, $N_r>N$, lead to worse approximations. 

In summary, our results show that also the \NSGAT can compute constant-factor approximations of the Pareto front. Different from the \NSGA (with sequential survival selection or in the steady-state mode), the absolute population size is less important (in particular, increasing the population size does not give better approximations), but the relation to the number of reference points is important for the approximation ability of the \NSGAT. In particular, we observe that for approximating the Pareto front, it appears best that $N_r$ and $N$ are very close, in contrast to the existing result for computing the full Pareto front~\cite{WiethegerD23,OprisDNS24}, which all require $N$ to be at least a constant factor larger than $N_r$. 

\section{Preliminaries}

This section will give a brief introduction on the \NSGAT, the algorithm to analyze, \omm, the benchmark to optimize, and the approximation metric that we will use.

\subsection{NSGA-III}

The overall framework of the \NSGA is presented in Algorithm~\ref{alg:NSGA-III}. The \NSGAT, a variant of the \NSGA designed for many objectives, was proposed by Deb and Jain~\shortcite{DebJ14}, and its first runtime analysis was conducted by Wietheger and Doerr~\shortcite{WiethegerD23}. Same as the \NSGA, the \NSGAT maintains a population $P_t$ of a fixed size $N$ and generates an offspring population $Q_t$ of the same size in each iteration. Also the \NSGAT will remove $N$ individuals from the combined population $R_t=P_t\cup Q_t$ and first uses the non-dominated sorting~\cite{DebPAM02} to divide $R_t$ into $F_1, F_2, \dots$. The \NSGAT only differs in the secondary criterion used in the critical front $F_{i^*}$ for the survival selection. 
Instead of the crowding distance in the \NSGA, the \NSGAT uses the following reference point mechanism.
\begin{algorithm}[tb]
	\caption{NSGA-III}
	\label{alg:NSGA-III}
	\begin{algorithmic}[1]
		\STATE Let the initial population $P_0$ be composed of $N$ individuals chosen independently and uniformly at random from $\{0,1\}^n$
		\FOR {$t=0,1,2,\dots,$} 
		\STATE Generate the offspring population $Q_t$ with size $N$ \label{stp:offcla}
		\STATE Use fast-non-dominated-sort()~\cite{DebPAM02} to divide $R_t=P_t\cup Q_t$ into $F_1,F_2,\dots$
		\STATE Find $i^* \geq 1$ such that $\sum_{i=1}^{i^*-1}|F_i|<N$ and $\sum_{i=1}^{i^*}|F_i| \geq N$ \label{stp:Fi*}
		\STATE $Z_t = \bigcup_{i=1}^{i^*-1}F_i$ 
		\STATE Use 
        Algorithm~\ref{alg:selection} to select $\tilde{F}_{i^*}\subseteq F_{i^*}$ such that $|Z_t\cup \tilde{F}_{i^*}|=N$
		\STATE $P_{t+1} = Z_t \cup \tilde{F}_{i^*}$
		\ENDFOR
	\end{algorithmic}
\end{algorithm}

Initially, the objective function values are normalized, and subsequently, the normalized individuals are associated with reference points for selection. We adopted the structured reference point placement method used in~\cite{DebJ14}, which is based on the systematic approach by ~\cite{das1998normal}. In this approach, these reference points are uniformly distributed on the normalized hyperplane, specifically on a $(M-1)$-dimensional simplex, where each axis intercept is set to $1$. The total number of reference points $N_r$ for an $M$-objective problem with $p$ divisions along each objective is given by $N_r=\binom{M+p-1}{p}$. Figure~\ref{fig:ref} shows an example with $M = 3$ and $p = 4$. For bi-objective problems, we have $N_r=p+1$.
Due to the evenly distributed nature of reference points, the ideal scenario is for different individuals in the offspring population $P_{t+1}$ to be associated with different reference points. The greater the number and the more uniform the distribution of reference points associated with individuals in $P_{t+1}$ , the higher its diversity. In the selection of the critical front $F_{i^*}$ of the \NSGAT, the number of individuals already selected to $P_{t+1}$ and associated with a reference point $r$ is referred to as the niche count $\rho_r$ of that reference point. The algorithm prioritizes selecting individuals associated with the reference point that has the smallest niche count $\rho$ if such individuals exist. This strategy increases the number of reference points associated with individuals in $P_{t+1}$, thereby enhancing the diversity of the population. Algorithms~\ref{alg:normalization} and~\ref{alg:selection} respectively present the algorithmic frameworks for the normalization and selection processes. For further details, please refer to ~\cite{DebJ14,BlankDR19,WiethegerD23}.
\begin{figure}[tb]
	\centering
	\includegraphics[trim=0cm 4.5cm 0cm 2.5cm, clip,width=0.4\textwidth]{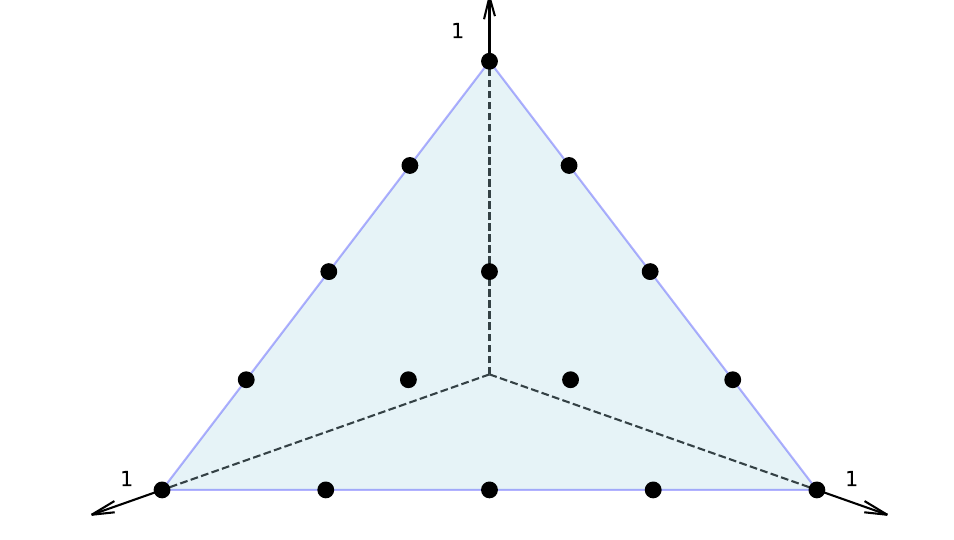}
	\caption{Structured reference points set for a three-objective problem with p = 4~\protect\cite{DebJ14}.}
	\label{fig:ref}
\end{figure}

\begin{algorithm}[tb]
	\caption{Normalization}
	\label{alg:normalization}
	\begin{algorithmic}[1]
		\REQUIRE $F_1,\dots,F_{i^*}$: non-dominated fronts
            \item[]  \hspace{\algorithmicindent} \space $f=(f_1,...,f_M)$: objective function
		\item[]  \hspace{\algorithmicindent} \space $z^*\in \R^M$: observed $\min$. in each objective
		\item[]  \hspace{\algorithmicindent} \space $z^w\in \R^M$: observed $\max$. in each objective
		\item[]  \hspace{\algorithmicindent} \space $E\subseteq\R^M$: extreme points of previous iteration, initially $\{\infty\}^M$
		\FOR {$j=1$ \textbf{to} $M$}  
		\STATE $\hat{z}_j^*=\min\{z_j^*,\min_{z\in Z}f_j(z)\}$
		\STATE Determine an extreme point $e^{(j)}$ in the $j$th objective from $R\cup E$ using an achievement scalarization function
		\ENDFOR
		\STATE $valid =$ False																					
		\IF {$e^{(1)},...,e^{(M)}$ are linearly independently}
		\STATE $valid =$ Ture
		\STATE Let $H$ be the hyperplane spanned by $e^{(1)},...,e^{(M)}$
		\FOR {$j=1$ \textbf{to} $M$}
		\STATE $I_j=$ the intercept of $H$ with the $j$th objective axis
		\IF {$I_j\ge\epsilon_{\nad}$ and $I_j\le z_j^w$}
		\STATE $\hat{z}_j^{\nad} = I_j$ \label{line:znad}
		\ELSE
		\STATE $valid =$ False
		\STATE \textbf{break}
		\ENDIF
		\ENDFOR
		\ENDIF 
		\IF {$valid =$ False}
		\FOR {$j=1$ \textbf{to} $M$}
		\STATE $\hat{z}_j^{\nad}=\max_{x\in F_1}f_j(x)$
		\ENDFOR
		\ENDIF
		\FOR {$j=1$ \textbf{to} $M$}
		\IF {$\hat{z}_j^{\nad}<\hat{z}_j^*+\epsilon_{\nad}$}
		\STATE $\hat{z}_j^{\nad}=\max_{x\in F_1\cup\dots\cup F_{i^*}}f_j(x)$
		\ENDIF
		\ENDFOR
		\STATE Define $f_j^n(x)=(f_j(x)-\hat{z}_j^{*})/(\hat{z}_j^{\nad}-\hat{z}_j^{*})$  \space $\forall x\in \{0,1\}^n, j\in \{1,...,M\}$
	\end{algorithmic}
\end{algorithm}

\begin{algorithm}[tb]
	\caption{Selection based on a set $U$ of reference points when maximizing the function $f$}
	\label{alg:selection}
	\begin{algorithmic}[1] 
		\REQUIRE $Z_t$: the multi-set of already selected individuals
		\item[]  \hspace{\algorithmicindent}  \space\space  $F_i^*$: the multi-set of individuals to choose from
		\STATE $f_n = \text{Normalize}(f, Z=Z_t \cup F_i^*)$ using Algorithm~\ref{alg:normalization}
		\STATE Associate each individual $x \in Z_t \cup F_i^*$ to the reference point rp$(x)$
		\STATE For each reference point $r \in U$, let $\rho_r$ denote the number of (already selected) individuals in $Z_t$ associated with $r$ \label{stp:niche}
		\STATE $U' = U$
		\STATE $\tilde{F}_{i^*} = \emptyset$
		\WHILE{True}
		\STATE Let $r_{\min} \in U'$ be such that $\rho_{r_{\min}}$ is minimal (break ties randomly)
		\STATE Let $x_{r_{\min}} \in F_{i^*} \setminus \tilde{F}_{i^*}$ be the individual that is associated with $r_{\min}$ and minimizes the distance between $f_n(x_{r_{\min}})$ and $r_{\min}$ (break ties randomly)
		\IF{$x_{r_{\min}}$ exists}
		\STATE $\tilde{F}_{i^*} = \tilde{F}_{i^*} \cup \{x_{r_{\min}}\}$
		\STATE $\rho_{r_{\min}} = \rho_{r_{\min}} + 1$ \label{stp:rhoplus}
		\IF{$|Z_t| + |\tilde{F}_{i^*}| = N$}
		\STATE \textbf{break all} and \textbf{return} $\tilde{F}_{i^*}$
		\ENDIF
		\ELSE
		\STATE $U' = U' \setminus \{r\}$
		\ENDIF
		\ENDWHILE
	\end{algorithmic}
\end{algorithm}

\subsection{\omm and Approximation Metric}
As mentioned before, the bi-objective \omm~\cite{GielL10} is the only benchmark used in the theoretical community to analyze the approximation ability of the \NSGA-type algorithms~\cite{ZhengD22gecco,ZhengD24approx}. For a good comparison, this work also chooses this bi-objective benchmark (and also the approximation metric used in these works) to analyze the approximation ability of the \NSGAT. We believe the techniques and the insights obtained in this work will be useful for the analysis for many objectives.



The \omm function is defined as follows, and we consider its maximization.
\begin{definition}[\cite{GielL10}]
For all search points $x$ the objective function $f:\{0,1\}^n\rightarrow \N \times \N$ is defined by
\begin{align*}
f(x)=(f_1(x),f_2(x))=\left(\sum_{i=1}^n x_i, n-\sum_{i=1}^n x_i\right).
\end{align*}
\end{definition}

In the language of multi-objective optimization, we say that $x$ \textit{weakly dominates} $y$, denoted as $x\succeq y$, if $f_1(x)\ge f_1(y)$ and $f_2(x)\ge f_2(y)$. If at least one of the two inequalities is strict, we say that $x$ \textit{dominates} $y$, denoted as $x\succ y$. If $x$ is not dominated by any individual in $\{0,1\}^n$, it is called \textit{Pareto optimal}, and the set of function values corresponding to all Pareto optimal solutions forms the \textit{Pareto front}. It is easy to see that the Pareto front of \omm is $\{(0,n),(1,n-1),\dots, (n,0)\}$, with a size of $n+1$.

As in the \NSGA's approximation theory~\cite{ZhengD22gecco,ZhengD24approx}, we will use the maximum empty interval (\mei) metric (See Definition~\ref{def:mei}) to evaluate how the algorithm approximates the Pareto front w.r.t. \omm. Note that the \mei can be easily transferred to other commonly used approximation metrics (like $\epsilon$-dominance~\cite{LaumannsTDZ02} and Hypervolume~\cite{ZitzlerT98})~\cite{ZhengD24approx}. 
\begin{definition}[\cite{ZhengD22gecco,ZhengD24approx}]\label{def:mei}
Let $S=\{(s_1,n-s_1), \dots, (s_m, n-s_m)\}$ be a subset of Pareto front $M$ of \oneminmax. Let $j_1,j_2,\dots,j_m$ be the sorted list of $s_1,\dots, s_m$ in the increasing order (ties broken uniformly at random). We define the maximum empty interval size of $S$, denoted by $\mei(S)$, as
\begin{align*}
\mei(S) = \max\{ j_1, n-j_m, j_{i+1}-j_i\mid \\i=1,\dots, m-1\}.
\end{align*}
For $n\in \N_{\ge 2}:=\{i\ge 2\mid i\in \N\}$, we further define
\begin{align*}
	\mei_{\opt}(N):=\min\{\mei(S)\mid &S\subseteq M,|S|\le N,\\&(0,n)\in S,(n,0)\in S\}.
\end{align*}
Obviously, this is the smallest \mei that an MOEA with a fixed population size $N$ can obtain when the extremal points $(0, n)$ and $(n, 0)$ are covered.
\end{definition}
The optimal $\mei$ value is shown in the following. Our work will obtain the upper bound of the \mei that the \NSGAT will achieve w.r.t. the \omm.
\begin{lemma}[\cite{ZhengD22gecco,ZhengD24approx}]\label{lem:meiopt}
For all $N\in \N_{\ge 2}$, we have $\mei_{\opt}(N) = \lceil\frac{n}{N-1}\rceil$.
\end{lemma}

\section{Approximation Guarantee when $N_r\le N$}\label{sec:app}
In this section, we will theoretically prove that when the number of reference points $N_r \leq N$, the \NSGAT can effectively approximate the Pareto front of the \omm problem.

\subsection{Reach Extremal Points}
To ease the discussion, here we call any optimal solution of one objective in a multiobjective problem an \emph{extremal point}. Note that it is not the same as the extreme point/vector used to determine the hyperpane, see in Algorithm~\ref{alg:normalization}. For \omm, $0^n$ and $1^n$ are the only extremal points. We will consider whether the two extremal points $0^n$ and $1^n$ can be reached. 

To achieve this, the following lemma will first show that for $N_r\le N$, the maximal and minimal objective values in the combined population survives. The key point is that after the normalization, such values will be mapped to $(1,0)$ or $(0,1)$. Note that $(1,0)$ and $(0,1)$ are reference points. Hence, the above values is closest to such reference points. Since $N_r\le N$ and all solutions are in $F_1$, we know that to select $N$ individuals in the survival selection in Algorithm~\ref{alg:selection}, the final $\rho_{r_{\min}}$ must be greater than or equal to $1$. Thus, before $\rho_{r_{\min}} > 1$, at least one individual with the maximal or minimal objective value (normalized to $(1,0)$ or $(0,1)$) will be selected into the next generation. Due to the pages limit, all proofs were moved to the supplementary material.

\begin{lemma}\label{lem:mmsave}
Consider using the \NSGAT with population size $N$ to optimize \omm with problem size $n$. Let $N < n + 1$ and $N_r\le N$. Let $z_1^{\min}:=\min\{f_1(x)\mid x\in R_t\}$ and $z_1^{\max}:=\max\{f_1(x)\mid x\in R_t\}$. Then $P_{t+1}$ will contain two individuals $x,y$ such that $f_1(x)=z_1^{\min}$ and $f_1(y)=z_1^{\max}$.
\end{lemma}

Note that the offspring generation operator (see Step~\ref{stp:offcla} in Algorithm~\ref{alg:NSGA-III}) will not increase $z_1^{\min}$ or decrease $z_1^{\max}$.
Then Lemma~\ref{lem:mmsave} shows that the minimal value of any objective will not increase and the maximal value of any objective will not decrease for all generations. Hence, the extremal points $0^n$ and $1^n$ will not be removed once they are reached. Thus, we focus on the time to decrease the minimal value of $f_1$ to $0$ to reach $0^n$, and the time to increase the maximal value of $f_1$ to $n$ to reach $1^n$. Lemma~\ref{lem:twoextreme} provides an upper bound on the runtime for the population to cover these two extremal points.

\begin{lemma}\label{lem:twoextreme}
Consider using the \NSGAT with population size $N$ to optimize \omm with problem size $n$. Let $N < n + 1$ and $N_r\le N$. Then within an expected number of $O(n \log n)$ iterations, that is, an expected number of $O(N n \log n)$ fitness evaluations, the two extremal points $0^n$ and $1^n$ will be reached for the first time, and both points will be kept in all future populations.
\end{lemma}

\subsection{Good Approximation Guarantee}
Lemma~\ref{lem:twoextreme} shows that the population will always contain $0^n$ and $1^n$ once they are reached for the first time. From basic calculations based on Algorithm~\ref{alg:normalization}, we obtain the following clear form for the normalized function value when the population contains both $0^n$ and $1^n$.

\begin{lemma}\label{lem:norm}
Consider using the \NSGAT with population size $N$ to optimize \omm with problem size $n$. Let $N < n + 1$ and $N_r\le N$. Assume that the two extremal points $0^n$ and $1^n$ are included in the population. Then for any individual $x$ in the population, its function value $f(x)$ is normalized to $f^n(x) = \frac{1}{n}f(x)$.
\end{lemma}
 

With the clear form of the normalized function value obtained in Lemma~\ref{lem:norm}, we easily mapping the function values to the normalized space so that we obtain the upper bound of the distance between the normalized function value and the associated reference point in Lemma~\ref{lem:dist} below.
\begin{lemma}\label{lem:dist}
Consider using the \NSGAT with population size $N$ to optimize \omm with problem size $n$. Let $N < n + 1$, $N_r\le N$, and let individual $x$ be associated with reference point $r=(r_1,r_2)$. Assume that the two extremal points $0^n$ and $1^n$ are included in the population. Normalize $f(x)$ to $f^n(x)=(f_1^n(x),f_2^n(x))$.
Then $f^n(x)$ is located in the non-negative region of the reference point plane and $|f_1^n(x) - r_1| \leq \frac{2 - \sqrt{2}}{N_r-1}$.
\end{lemma}
Since $0^n$ and $1^n$ are included in the population, Lemma~\ref{lem:dist} indicates a good mapping between the evenly distributed reference points $N_r$ and the desired good approximation of the evenly distributed function values. It will be our key proof idea for our approximation guarantee in Theorem~\ref{thm:runtime}. 

Note that not all reference points has associated individuals to be active in the survival selection. We need to analyze whether the evenly distributed references actually take their effect. For this, we first show in Lemma~\ref{lem:fgen} that once a reference point has an associated individual, it will contain at least one associated individual in all future generation. That is, once a reference point is active, it will be active forever. The key to the proof is that in the Algorithm~\ref{alg:selection}, each reference point will be selected at least once. Hence, at least one individual associated with the active reference point is retained.

\begin{lemma}\label{lem:fgen}
Consider using the \NSGAT with population size $N$ to optimize \omm with problem size $n$. Let $N < n + 1$ and $N_r\le N$. Assume that the two extremal points $0^n$ and $1^n$ are included in the population. Then, if a reference point $r$ has individuals associated with it, there will always be at least one individual associated with it in future generations.
\end{lemma}


As the active reference point will remain active as shown in Lemma~\ref{lem:fgen}, we now calculate the time to activate all reference points, that is, to reach a status when all reference points have their associated individuals. From Lemma~\ref{lem:dist} we have the estimate between the active reference point and its associated individual's normalized function value. Then we could use the waiting time argument to estimate the time from an active reference point to generate a neighbor non-activated reference points. Hence, the overall time for activating all reference points is obtained in Lemma~\ref{lem:runtime}.

\begin{lemma}\label{lem:runtime}
Consider using the \NSGAT with population size $N$ to optimize \omm with problem size $n$. Let $N < n + 1$ and $N_r\le N$. Let $c:=\lceil\frac{2(2-\sqrt{2})n}{N_r-1}\rceil$. Assume that the two extremal points $0^n$ and $1^n$ are included in the population. Then after an expected number of $O(n^c\log n)$ iterations, that is, an expected number of $O(Nn^c\log n)$ fitness evaluations, each reference point $r$ has at least one individual $x$ associated with it and this property will be kept for all future time.
\end{lemma}

Although Lemma~\ref{lem:runtime} estimates the runtime for activating all reference points, the associated individuals do not must have its normalized function value same as its associated reference point. Hence, we now use Lemma~\ref{lem:dist} to estimate the distance between individuals associated to neighbor reference points, and then obtain an upper bound on \mei (see the following lemma).
\begin{lemma}\label{lem:upb}
Consider using the \NSGAT with population size $N$ to optimize \omm with problem size $n$. Let $N < n + 1$, $N_r\le N$, and let $t_0$ be the first generation such that the two extremal points $0^n$and $1^n$ are included in the population. Assume that each reference point $r$ has at least one individual $x$ associated with it. Then for any $t \geq t_0$, we have $\mei\leq \lceil \frac{(5 - 2\sqrt{2})n}{N_r-1}\rceil$.
\end{lemma}


Hence, from Lemmas~\ref{lem:twoextreme},~\ref{lem:runtime} and~\ref{lem:upb}, we obtain the following theorem regarding the approximation ability of \NSGAT.

\begin{theorem}\label{thm:runtime}
Consider using the \NSGAT with population size $N$ to optimize \omm with problem size $n$. Let $N < n + 1$ and $N_r\le N$. Let $c:=\lceil\frac{2(2-\sqrt{2})n}{N_r-1}\rceil$. Then after an expected number of $O(Nn^c\log n)$ fitness evaluations, a population containing the two extremal points $0^n$ and $1^n$ and with $\mei\le\lceil\frac{(5-2\sqrt{2})n}{N_r-1}\rceil$ is reached and both properties will be kept for all future time.
\end{theorem}
From Theorem~\ref{thm:runtime}, we see both the upper bound of the $\mei$ value and the runtime to reach such upper bound heavily depend on $N_r$, the number of reference points. It is quite different from the \NSGA (with current crowding distance) whose upper bound $\mei\le \max\{\frac{2n}{N-3},1\}$~\cite{ZhengD24approx} merely relies on $N$, the population size. Note that the optimal $\mei=\lceil \frac{n}{N-1}\rceil$ merely depends on $N$, the number of points, as well. Besides, Theorem~\ref{thm:runtime} shows that the minimal $\mei$ upper bound of $\lceil\frac{(5-2\sqrt{2})n}{N-1}\rceil$ that the \NSGAT can reach happens when $N_r=N$. Although this upper bound is slightly worse than the one for the \NSGA with current crowding distance discussed before, our experiments in Section~\ref{sec:exp} shows the \NSGAT always reaches the optimal \mei, and performs better than the \NSGA.

\section{Possible Difficulty for Large $N_r>N$}\label{sec:largenr}
In contrast to the approximation guarantee in the previous section, this section will show  possible difficulties of the \NSGAT with large enough $N_r$. It is quite different from the existing theoretical results~\cite{WiethegerD23,OprisDNS24} that require the large enough $N_r$ for their analyses.

\subsection{Lose Extremal Points}
Recall the definition of the extremal pointed in the above section. For \omm, $0^n$ and $1^n$ are the only extremal points. Extremal points usually carry the special information and are expected to be maintained. In the \NSGA, for the multiset of any extremal point, at least one of the repetitions will have the infinite crowding distance and will survive to the next generation if the population size is large enough, say at least $8$ for \omm. However, the \NSGAT has no such a special treatment to ensure the extremal points survive to the next generation. Instead, the removal is determined by the numbers of already chosen individuals associated to the reference points, and the extremal points have the chance to be removed. The following lemma gives an example where the extremal point can be removed.

\begin{lemma}
Consider using the \NSGAT with population size $N$ to optimize \omm with problem size $n$. Let $N \le (n+1)/2$ and $N_r$ be large enough. Assume that before the survival selection, in the combined parent and offspring population $R_t$ that contains $0^n$ and $1^n$, each reference point has at most one associated individual. Then after the survival selection, the probabilities to remove one specific extremal point, to remove at least one specific extremal point, and to remove both extremal points are $\frac12, \frac34+\frac{1}{4(2N-1)},$ and $\frac14 - \frac{1}{4(2N-1)}$ respectively.
\label{lem:lose}
\end{lemma}

The above lemma shows that with a good probability, the survival selection will lose the extremal points for this example.

\subsection{Possible Bad Approximation Quality}
In addition to the above example that the extremal points can be removed, here we discuss an example where the survival selection of the \NSGAT can lead to a quite bad approximation from the optimal-looking situation. This situation is essentially similar to the one used to demonstrate the possible difficulty of the traditional \NSGA in~\cite{ZhengD24approx}. 

\begin{lemma}
    Consider using the \NSGAT with problem size $N$ to optimize the \omm with problem size $n$. Let $n$ be odd and $N=(n+1)/2$ and $N_r=2N$. Assume that the combined parent and offspring population $R_t$ covers the full Pareto front. Then the next population has the expected $\mei$ value of $\Omega(\log n)$.
\end{lemma}
The optimal $\mei$ value of $N=(n+1)/2$ is $\lceil \frac{n}{N-1}\rceil \le 4$. The above lemma shows that even from the above optimal-looking combined population $R_t$, the $\mei$ value becomes $\Omega(\log n)$ in expectation. Due to the complicated process of the \NSGAT, we currently do not know how often the above situation happens. However, our experiments in Section~\ref{sec:exp} show that the difficulties of the \NSGAT with large number of reference points exist indeed.

As mentioned before, this work focuses on the approximation ability of the original \NSGAT. Hence, here we will not discuss the strategies to overcome the above difficulty stemming from losing extremal points, but leave it as our interesting future research.

\section{Experiments}\label{sec:exp}
Previous sections gave the theoretical approximation results for the \NSGAT on \omm. This section conducts the extensive experiments to address the following three concerns.
\begin{itemize}
    \item \textbf{The impact of the number of reference points $N_r$.} Our theorem shows that the best approximation ability of the \NSGAT on \omm happens when $N_r=N$. We will see whether it is experimentally true for the \omm. Besides, Section~\ref{sec:largenr} proved the difficulties of the \NSGAT with $N_r>N$ on special cases. It remains unknown whether it actually happens.
    \item \textbf{The actual approximation ability.} The upper bound of the \mei approximation metric obtained in Theorem~\ref{thm:runtime} for the \omm is $5-2\sqrt 2$ factor larger than the optimal value, and slightly worse than the theoretical bound for the \NSGA (with sequential survival selection). We will see how well the \NSGAT experimentally approximates, together with its comparison with the \NSGA.
    \item \textbf{The verification of the above findings beyond \omm.} The above two concerns are for the verification on \omm as in our theoretical results. A natural question is about the behaviors of the \NSGAT on other problems. 
\end{itemize}

\subsection{Experimental Settings}
To address the first two concerns w.r.t. the \omm, we adopt the same settings in the only theoretical approximation works of the \NSGA~\cite{ZhengD22gecco,ZhengD24approx}. That is, we consider the problem size $n=601$ and population sizes $N=301,151,76$ ($N=\lceil (n+1)/2\rceil,\lceil (n+1)/4\rceil,\lceil (n+1)/8\rceil$) in these works for the easy comparison to the \NSGA (with sequential survival selection, denoted as \NSGA' in the following). We set the number of reference points $N_r=\lceil N/4\rceil,\lceil N/2\rceil,N,2N,4N,8N$, with the first three for $N_r\le N$ discussed in Section~\ref{sec:app} and the last three for $N_r>N$ discussed in Section~\ref{sec:largenr}.

To tackle the last concern, we experimentally consider another popular \lotz benchmark, and the optimal $\mei=\lceil \frac{n}{N-1}\rceil$. We set the problem size $n=120$, the setting used in~\cite{ZhengD23aij}, and the population size $N=\lceil (n+1)/2\rceil,\lceil (n+1)/4\rceil,\lceil (n+1)/8\rceil$ and the number of reference points $N_r=\lceil N/4\rceil,\lceil N/2\rceil,N,2N,4N,8N$, same settings as for \omm.

$20$ independent runs are conducted.
We regard these numbers are enough as the collected data are quite concentrated.

\subsection{Experimental Results}
Due to the limited space, here we only report the results for $N=\lceil (n+1)/8\rceil$ (that is, $N=76$ for the \omm and $N=16$ for the \lotz), and put the results for other settings of the population size into the supplementary material. Note that experimental findings obtained for this setting of $N$ also similarly hold for other $N$.

\begin{table}[tb]
\centering
\caption{The first, second, and third quartiles (displayed in the format of $(\cdot,\cdot,\cdot)$) for the maximal empty interval sizes \mei within 100 generations and 20 independent runs. Generations $[1..100]$ and $[3001..3100]$ after $T_{\start}$ on the \omm problem with $n=601$ and $N=76$ are regarded separately. Here, $T_{\start}$ is the generation number when the population contains both extremal points for the first time. For the  case of losing extremal points for $N_r>N$, it is set to be $T_{\max}$, the maximal value of $T_{\start}$ reported for the algorithm with the settings $N_r\le N$ in $20$ independent runs.}
\begin{tabular*}{0.4\textwidth}{@{\extracolsep{\fill}} lcc}
    \hline
    \addlinespace[2pt]
    Generations  & [1..100] & [3001..3100] \\
    \addlinespace[1pt]
    \hline
    \addlinespace[1pt]
    \NSGA'&(11,12,12)&(11,12,12)\\
    $N_r=\lceil N/4\rceil$&(33,33,33)&(33,33,33.25)\\
    $N_r=\lceil N/2\rceil$&(17,17,17)&(17,17,17)\\ 
    $N_r=N$&(9,9,9)&(9,9,9)\\ 
    $N_r=2N$&(187,194,203)&(186,194,203)\\ 
    $N_r=4N$&(229,237,244)&(236,240,245)\\ 
    $N_r=8N$&(228,234,240)&(229,240,248)\\
    \addlinespace[1pt]
    \hline
\end{tabular*}
\label{tab:meiomm}
\end{table}
\begin{figure}[tb]
	\centering
	\includegraphics[width=0.45\textwidth]{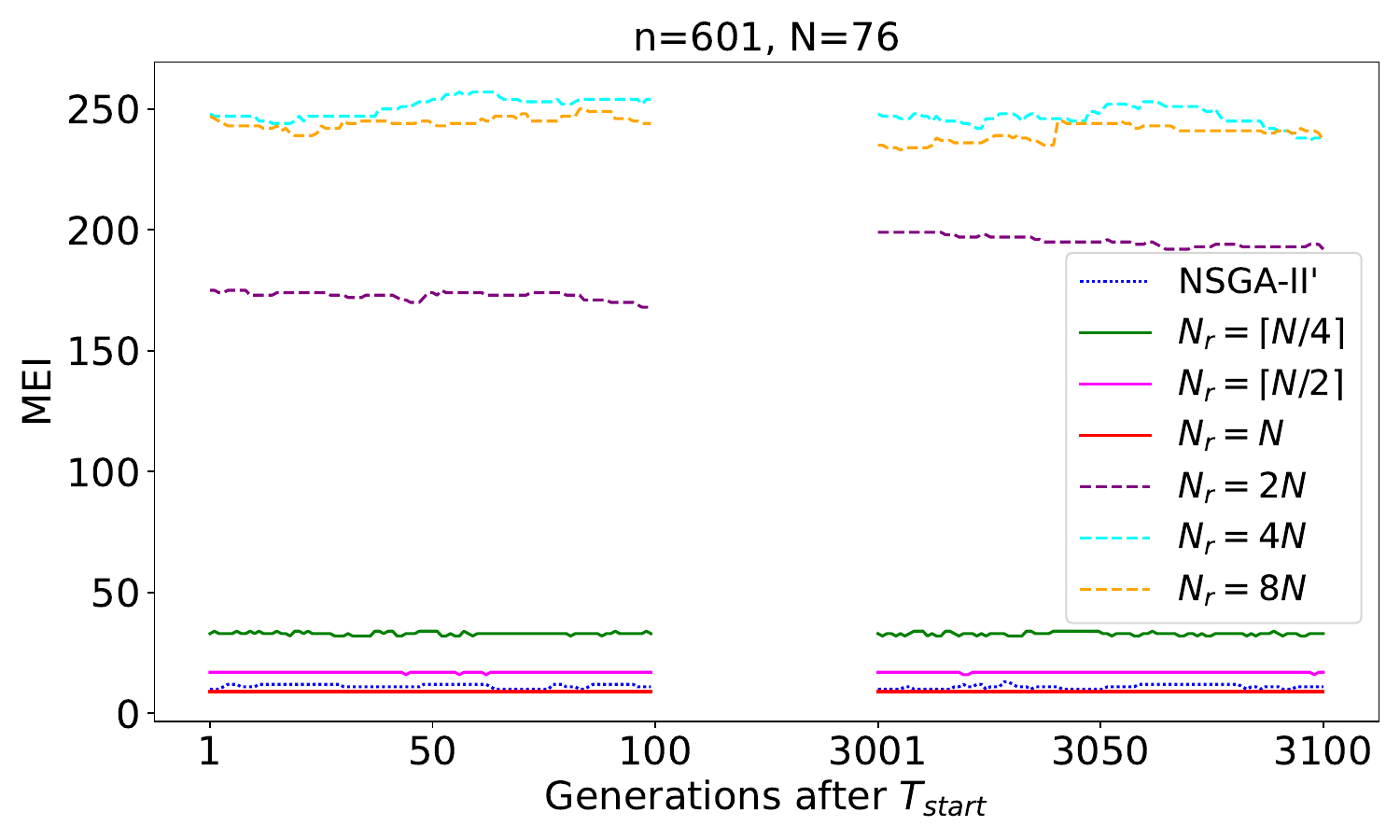}
	\caption{The \mei for generations $[1..100]$ and $[3001..3100]$ after $T_{\start}$, in one exemplary run on the \omm problem with $n=601$ and $N=76$.}
	\label{fig:meiomm}
\end{figure}
\subsubsection{The Impact of $N_r$} 
From the stable \mei values reported in Table~\ref{tab:meiomm} for 20 runs and Figure~\ref{fig:meiomm} for one exemplary run, we see that $N_r=N$ achieves the best \mei value, compared with other settings of $N_r$. We also see that small values of $N_r<N$ leads to significantly better approximation quality against large values of $N_r>N$.
%
Furthermore, in the experiments, we observed that the algorithm terminates only when $T_{\max}$ is reached, which shows that the difficulty to have both $0^n$ and $1^n$ exists and verifies the possible difficulty of the \NSGAT with large enough number of reference points discussed in Section~\ref{sec:largenr}.


\subsubsection{The Actual Approximation Ability}
Note that for $n=601$ and $N=76$ reported in Table~\ref{tab:meiomm}, the optimal $\mei=\lceil \frac{n}{N-1} \rceil=9$. Hence, for the best setting of $N_r=N$, we see all three quartiles reach this optimal value. Together with one exemplary one plotted in Figure~\ref{fig:meiomm} and all data in generations $[3001..3100]$), we know that the \NSGAT will reach the optimal \mei for \omm, while the \NSGA' (the sequential version of the \NSGA analyzed in~\cite{ZhengD24approx}) cannot. This result shows that the upper bound $\lceil\frac{(5-2\sqrt2)n}{N-1}\rceil$ (with $N_r=N$) obtained in Theorem~\ref{thm:runtime} is not tight, and we leave it as our interesting future research.
%

\subsubsection{The Approximation Ability of the \NSGAT on \lotz}
Analogous to Table~\ref{tab:meiomm} and Figure~\ref{fig:meiomm} for the \omm problem, Table~\ref{tab:meilotz16} (20 runs) and Figure~\ref{fig:meilotz16} (one exemplary run) are for the \lotz problem. We see that $N_r=N$ is the best setting for the approximation ability of the \NSGAT on \lotz, and this best setting results in the optimal \mei value of $8$ (note that the Pareto front for the \lotz is $\{(0,n),(1,n-1),\dots, (n,0)\}$, which is the same for the \omm. Hence, the optimal \mei value is $\mei=\lceil \frac{n}{N-1} \rceil=8$ for $n=120$ and $N=16$).

\begin{table}[tb]
	\centering
	\caption{The first, second, and third quartiles (displayed in the format of $(\cdot,\cdot,\cdot)$) for the maximal empty interval sizes \mei within 1000 generations after $T'_{\start}$ in 20 independent runs on the \lotz problem with $n=120$ and $N=16$. Here, $T'_{\start}$ is the generation number when the population contains both extremal points and has $\mei=\lceil \frac{n}{N_r-1}\rceil$ for the first time. For the case of losing extremal points for $N_r>N$, it is set to be $T'_{\max}$, the maximal value of $T'_{\start}$ reported for the algorithm with the settings $N_r\le N$ in $20$ independent runs. 
    }
\begin{tabular*}{0.4\textwidth}{@{\extracolsep{\fill}} lc}
	\hline
	\addlinespace[2pt]
	Generations  &After $T'_{\start}$\\
	\addlinespace[1pt]
	\hline
        \addlinespace[2pt]
        $N_r=\lceil N/4\rceil$&(40,40,40)\\
        $N_r=\lceil N/2\rceil$&(18,18,18)\\ 
	$N_r=N$&(8,8,8)\\ 
        $N_r=2N$&(19,22,24)\\
        $N_r=4N$&(41,50,61)\\
        $N_r=8N$&(45,52,63)\\
	\addlinespace[1pt]
	\hline
\end{tabular*}
\label{tab:meilotz16}
\end{table}
\begin{figure}[tb]
	\centering
	\includegraphics[width=0.45\textwidth]{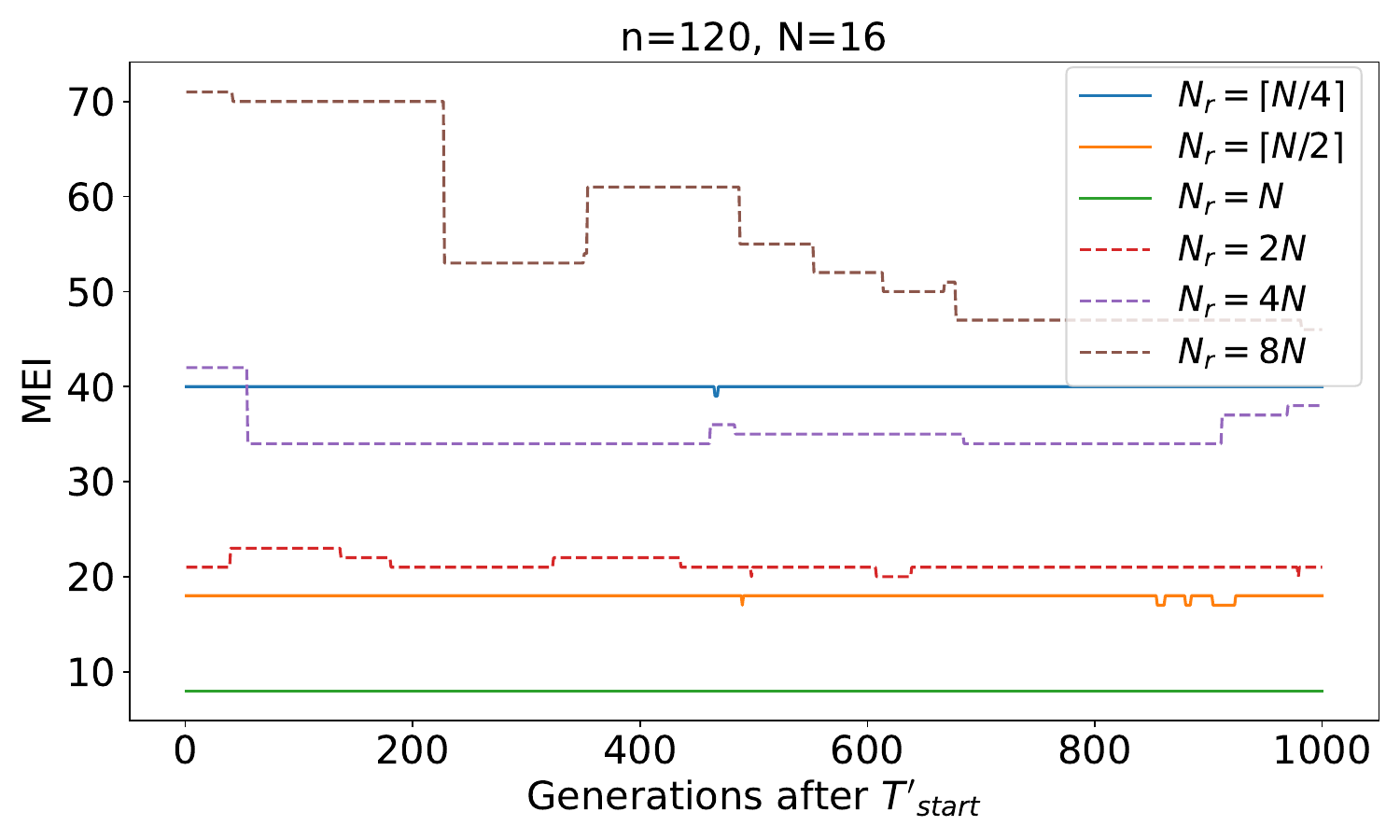}
	\caption{The \mei for generations $[1..1000]$ after $T'_{\start}$, in one exemplary run on the \lotz problem with $n=120$ and $N=16$.}
	\label{fig:meilotz16}
\end{figure}

\section{Conclusion}
This paper conducted the first theoretical analysis for the approximation performance of the \NSGAT. We showed that the number of reference points plays an essential role for the approximation quality of the \NSGAT, considering the \omm benchmark. 

In detail, we proved that when the population size $N$ is smaller than the size of the Pareto front, the \NSGAT can achieve an \mei approximation value of $\lceil\frac{(5-2\sqrt2)n}{N_r-1}\rceil$ on the \omm problem within an expected $O(Nn^c\log n)$ function evaluations, where $N_r\le N$ is the number of reference points and $c=\lceil\frac{2(2-\sqrt2)n}{N_r-1}\rceil$. Then the best upper bound of the approximation quality (only a constant factor of $5-2\sqrt2$ larger than the optimal $\mei$) is witnessed for $N_r= N$. However, when $N_r>N$, we proved the possible approximation difficulties of the \NSGAT for some examples. 

Our experiments verified the above findings, say the good approximation ability of the \NSGAT with $N_r\le N$ and its bad approximation ability for $N_r>N$. Our experiments also showed that the \NSGAT with $N_r=N$ can reach the optimal $\mei$ for approximation, which will guide us to derive a tighter approximation bound as our future interesting research topic.

We note the differences compared to the existing theoretical works. For the approximation ability of the \NSGA~\cite{ZhengD24approx}, the population size influences the upper bound of the $\mei$ value. However, there the number of reference points has the essential influence. Besides, different from the existing theory works~\cite{WiethegerD23,OprisDNS24}, a large enough number of the reference points are required to establish the runtime theory for the full coverage of the Pareto front. This work shows that a large number of reference points can be harmful for the approximation.

\section*{Acknowledgments}
This work was supported by National Natural Science Foundation of China (Grant No. 62306086, 62350710797), Science, Technology and Innovation Commission of Shenzhen Municipality (Grant No. GXWD20220818191018001), and Guangdong Basic and Applied Basic Research Foundation (Grant No. 2025A1515011936). This research benefited from the support of the FMJH Program PGMO.
}


\end{document}